\documentclass[conference]{IEEEtran}

\usepackage{caption}
\usepackage{subcaption}

\usepackage{graphics} % for pdf, bitmapped graphics files
\usepackage{epsfig} % for postscript graphics files

\usepackage[font=small]{caption}

\usepackage{times} % assumes new font selection scheme installed
\usepackage{xcolor} % allows colored text
\usepackage{units} % allows nice unit formatting
\usepackage{graphicx}
\usepackage{latexsym}
\usepackage{setspace}
\usepackage{multicol}
\usepackage{verbatim}
\usepackage{float}
\usepackage{cite}
\usepackage{subcaption}
\usepackage[export]{adjustbox}
\usepackage{soul}
\usepackage{stfloats}
\usepackage{algorithm} 
\usepackage{algcompatible}

\newcommand{\multiline}[1]{%
  \begin{tabularx}{\dimexpr\linewidth-\ALG@thistlm}[t]{@{}X@{}}
    #1
  \end{tabularx}
}
\usepackage{arydshln} % dashed line in table

\usepackage{amssymb,amsmath,bm}

\def\beq{\begin{equation}}
\def\eeq{\end{equation}}
\def\bql{\begin{equation}}
\def\eql{\end{equation}}
\def\bqn{\begin{eqnarray*}}
\def\eqn{\end{eqnarray*}}
\def\bnl{\begin{eqnarray}}
\def\enl{\end{eqnarray}}
\def\bna{\bql\begin{array}{rcl}}
\def\ena{\end{array}\eql}
\def\bnn{\beq\begin{array}{rcl}}
\def\enn{\end{array}\eeq}
\def\bma{\begin{bmatrix}}
\def\ema{\end{bmatrix}}
\def\bmx{\begin{matrix}}
\def\emx{\end{matrix}}
\def\ben{\begin{enumerate}}
\def\een{\end{enumerate}}
\def\bit{\begin{itemize}}
\def\eit{\end{itemize}}
\def\bei{\begin{itemize}}
\def\eei{\end{itemize}}
\def\bet{\begin{tabular}}
\def\eet{\end{tabular}}

\newcommand{\allcaps}[1]{\uppercase\expandafter{#1}}

\IEEEoverridecommandlockouts                              % This command is only needed if 
\begin{document}

\title{Design and Flight Demonstration of a Quadrotor for Urban Mapping and Target Tracking Research}

\author{
\IEEEauthorblockN{
Collin Hague\IEEEauthorrefmark{1}, 
Nicholas Kakavitsas\IEEEauthorrefmark{1}, 
Jincheng Zhang\IEEEauthorrefmark{2}, 
Chris Beam\IEEEauthorrefmark{2}, 
Andrew Willis\IEEEauthorrefmark{2},
and Artur Wolek\IEEEauthorrefmark{1}}
\IEEEauthorblockA{
\IEEEauthorrefmark{1}
Department of Mechanical Engineering and Engineering Science}
%University of North Carolina at Charlotte
%Charlotte, NC 28223 USA\\
\IEEEauthorblockA{
\IEEEauthorrefmark{2}
Department of Electrical and Computer Engineering\\
Email: \{ chague, nkakavit, jzhang72, cbeam18, arwillis, awolek\} @charlotte.edu \\
University of North Carolina at Charlotte, \\
Charlotte, NC 28223 USA}
}

%\author{\IEEEauthorblockN{Michael Shell}
%\IEEEauthorblockA{School of Electrical and\\Computer Engineering\\
%Georgia Institute of Technology\\
%Atlanta, Georgia 30332--0250\\
%Email: http://www.michaelshell.org/contact.html}
%\and
%\IEEEauthorblockN{Homer Simpson}
%\IEEEauthorblockA{Twentieth Century Fox\\
%Springfield, USA\\
%Email: homer@thesimpsons.com}
%\and
%\IEEEauthorblockN{James Kirk\\ and Montgomery Scott}
%\IEEEauthorblockA{Starfleet Academy\\
%San Francisco, California 96678--2391\\
%Telephone: (800) 555--1212\\
%Fax: (888) 555--1212}}
%\author{
%\IEEEauthorblockN{Jacob Herbert and Artur Wolek}
%\IEEEauthorblockA{Department of Mechanical Engineering and Engineering Science\\
%University of North Carolina at Charlotte\\
%Charlotte, NC 28223 USA \\
%Email: jherbert@charlotte.edu, awolek@charlotte.edu}
%}

\maketitle

\begin{abstract}
This paper describes the hardware design and flight demonstration of a small quadrotor with imaging sensors for urban mapping, hazard avoidance, and target tracking research.  
The vehicle is equipped with five cameras, including two pairs of fisheye stereo cameras that enable a nearly omnidirectional view and a two-axis gimbaled camera. An onboard NVIDIA Jetson Orin Nano computer running the Robot Operating System software is used for data collection. An autonomous tracking behavior was implemented to coordinate the motion of the quadrotor and gimbaled camera to track a moving GPS coordinate. The data collection system was demonstrated through a flight test that tracked a moving GPS-tagged vehicle through a series of roads and parking lots. A map of the environment was reconstructed from the collected images using the Direct Sparse Odometry (DSO) algorithm. The performance of the quadrotor was also characterized by acoustic noise, communication range, battery voltage in hover, and maximum speed tests.
\end{abstract}

\let\thefootnote\relax\footnotetext{Copyright © 2024 IEEE. Personal use of this material is permitted. Permission from IEEE must be obtained for all other uses, in any current or future media, including reprinting/republishing this material for advertising or promotional purposes, creating new collective works, for resale or redistribution to servers or lists, or reuse of any copyrighted component of this work in other works by sending a request to pubs-permissions@ieee.org.}

\section{Introduction}
Small multirotor uncrewed aerial vehicles (UAV) can be used for urban mapping and target tracking applications, such as tracking people or vehicles by law enforcement, aerial videography, crowd monitoring, disaster search and rescue, and military surveillance \cite{alcantara2020autonomous, schedl2021autonomous, geiger2008flight}. However, urban environments are cluttered areas, containing trees, power lines, buildings, birds, and other hazards that make navigation difficult. Onboard sensors and associated algorithms must be capable of performing the tracking task,  perceiving static and dynamic hazards, and simultaneously building maps of the environment. %offer a means for quadrotors to adjust their trajectories to avoid collisions. 
Hazard avoidance enables quadrotors to operate at low altitudes in urban areas, and mapping data products can be used to  georeference target tracks or support GPS-denied navigation. 

The selection of perception algorithms and onboard sensors is critical in enabling robust flight of UAVs in urban spaces. 
%The selection of sensors and their positioning on the platform is critical for developing robust perception capabilities that can operate in near real-time onboard UAVs. 
Real-time processing demands high-performance computing hardware, which impacts the size, weight, and power considerations of the vehicle. The anticipated operating environment also drives sensor selection---operating in varying lighting conditions such as day and night requires infrared cameras, LiDAR, and/or radar, while daytime mapping, tracking, and avoidance under ideal conditions may use exclusively electro-optic cameras. Perception algorithms are tailored for specific sensors and require tightly coupled integration with  control systems. Therefore, well-designed interfaces,  communication protocols, and complementary control and perceptional algorithm selection are essential for robust urban UAV flight.

\begin{figure}[t]
    \centering
    \includegraphics[height = 0.6 in]{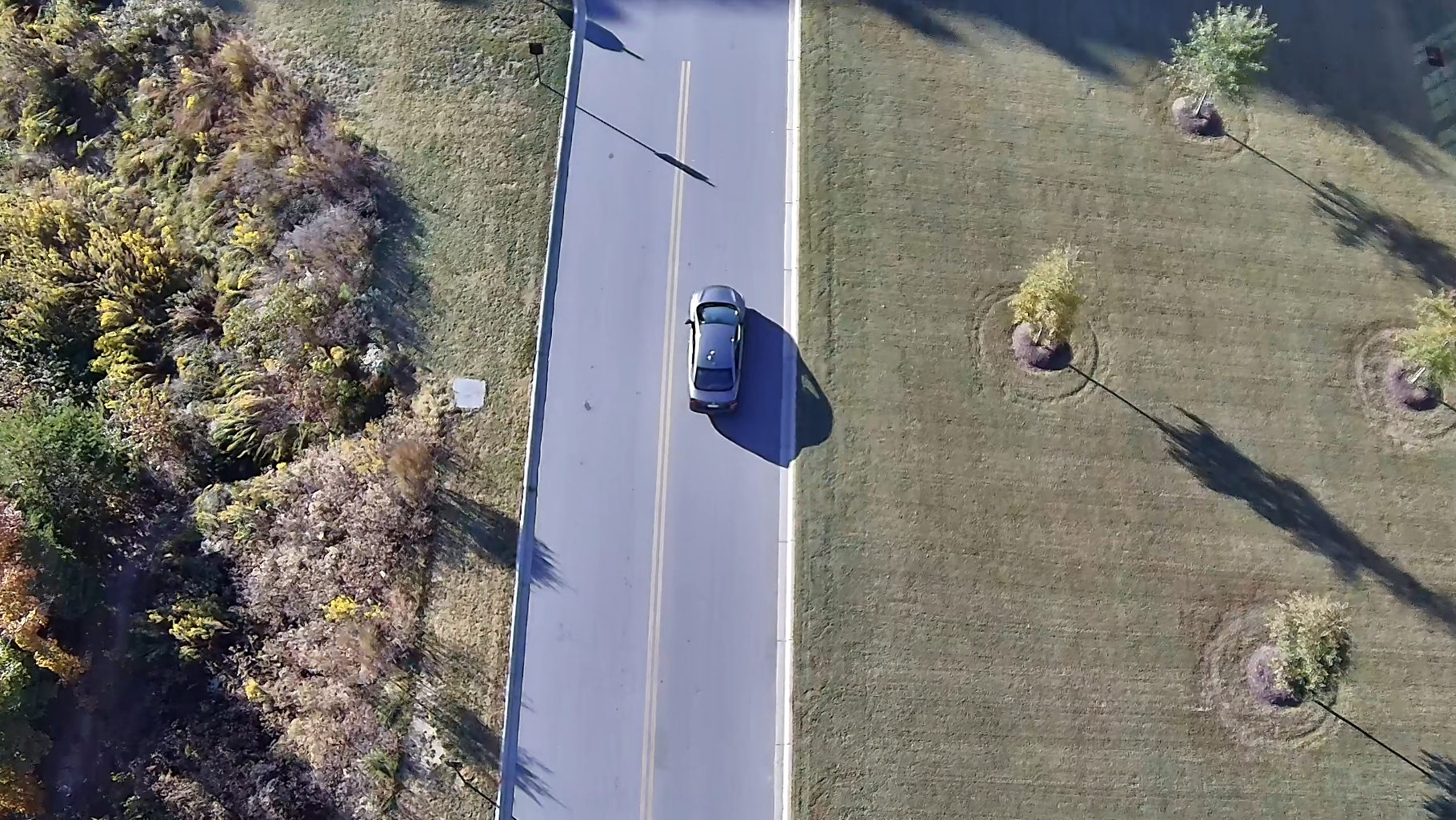}~
    \includegraphics[height = 0.6 in]{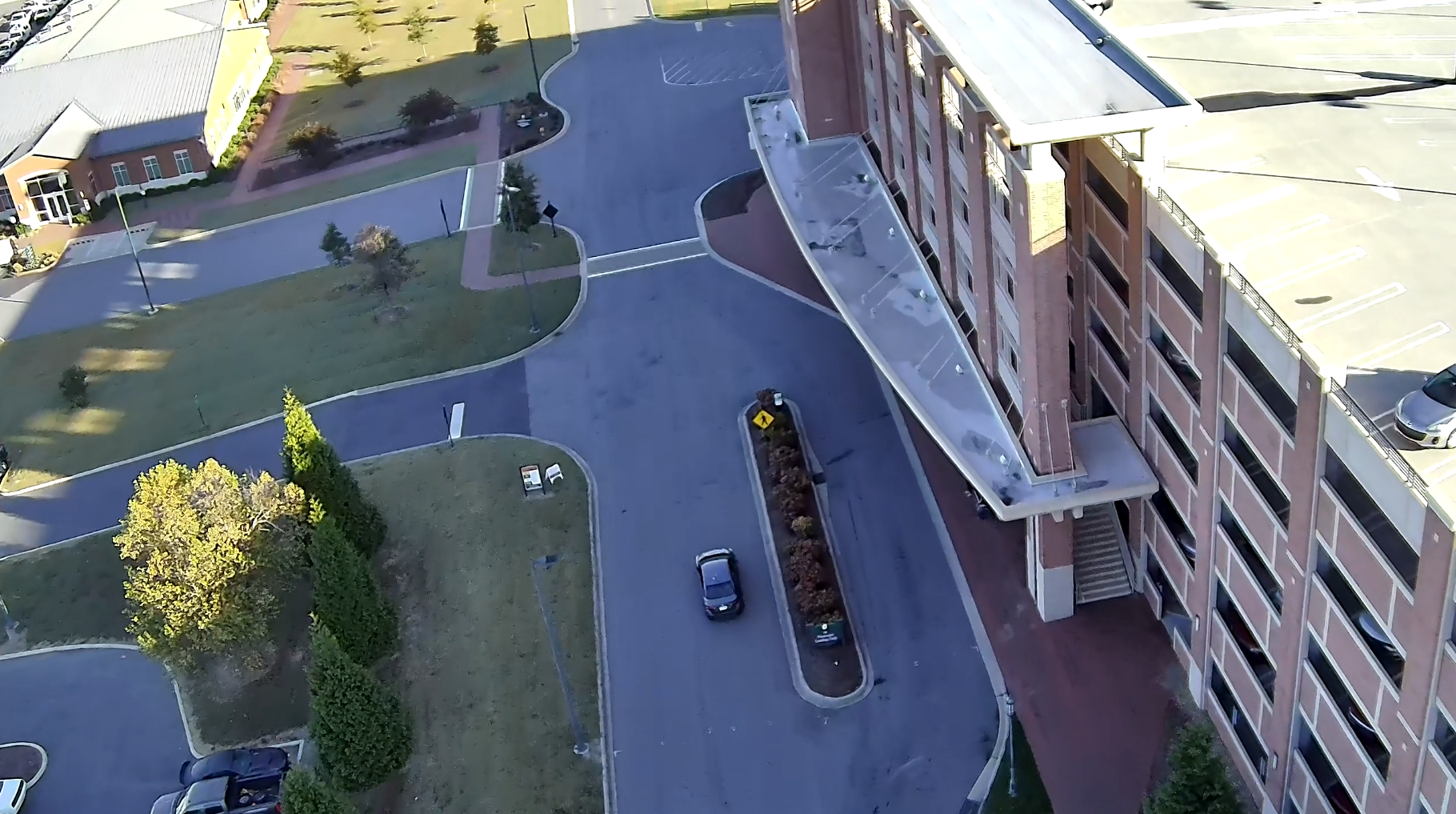}~
    \includegraphics[height = 0.6 in]{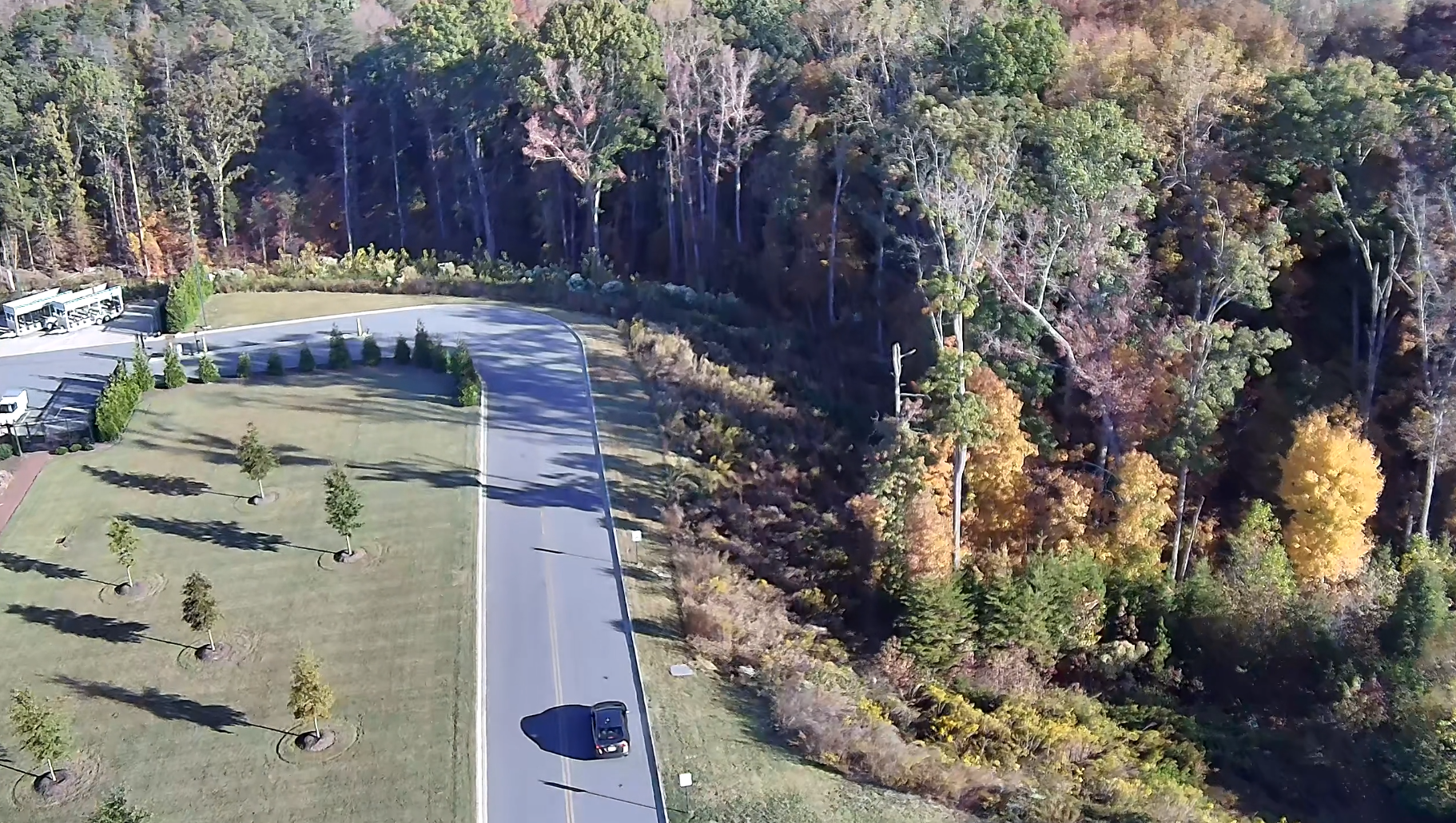}\\  \vspace{1em}
    \includegraphics[height = 0.6 in]{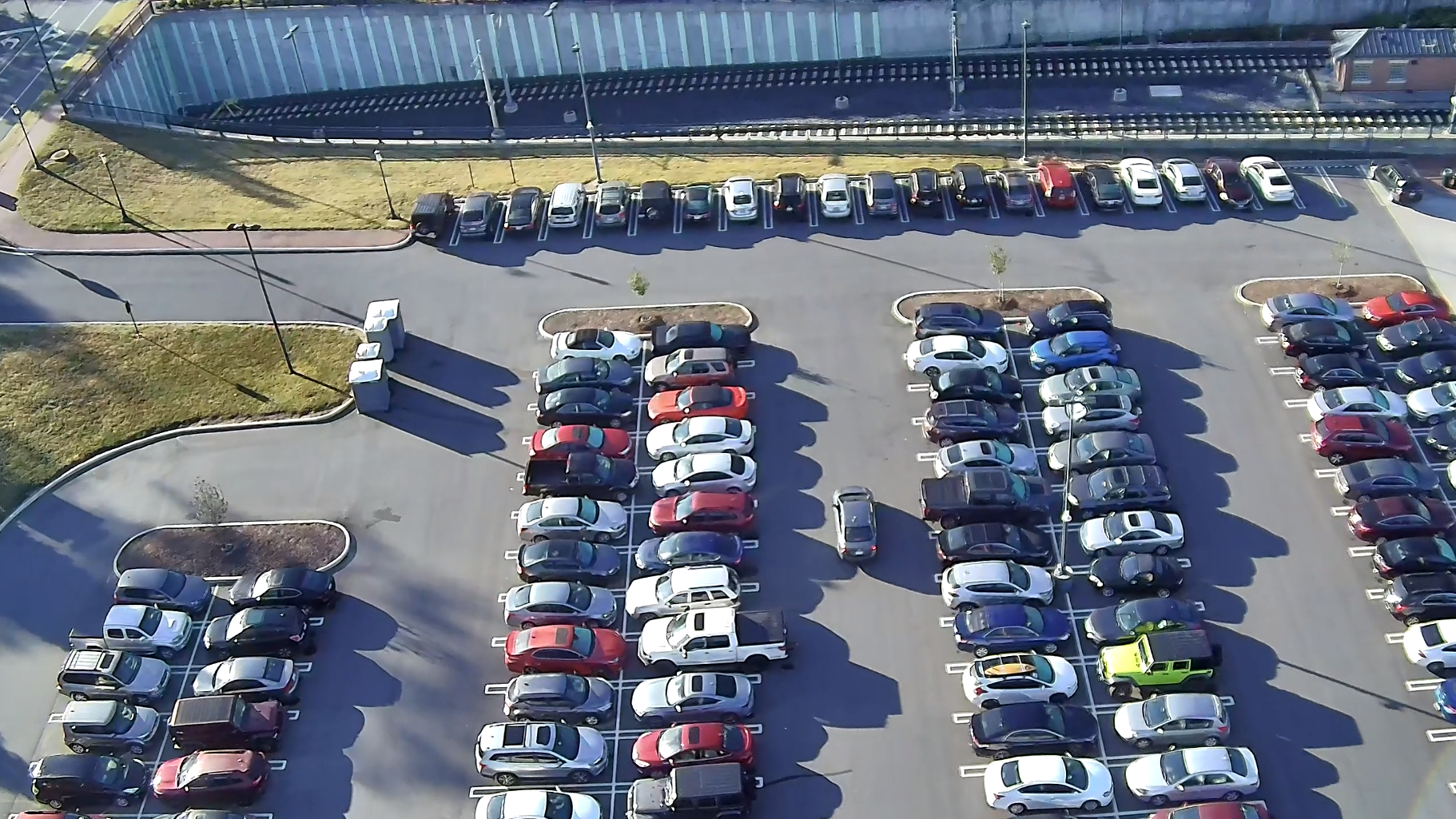}~
    \includegraphics[height = 0.6 in]{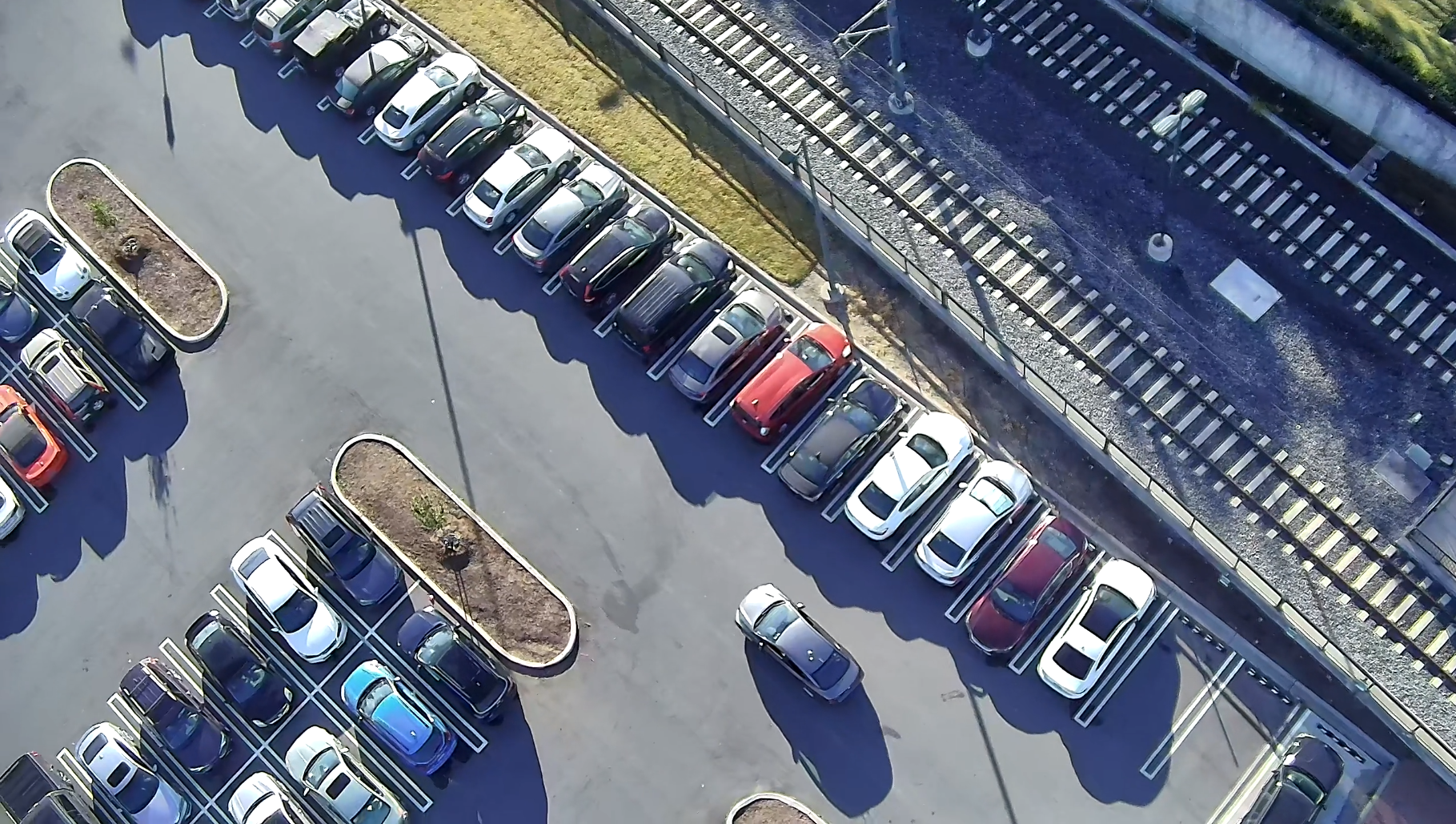}~
    \includegraphics[height = 0.6 in]{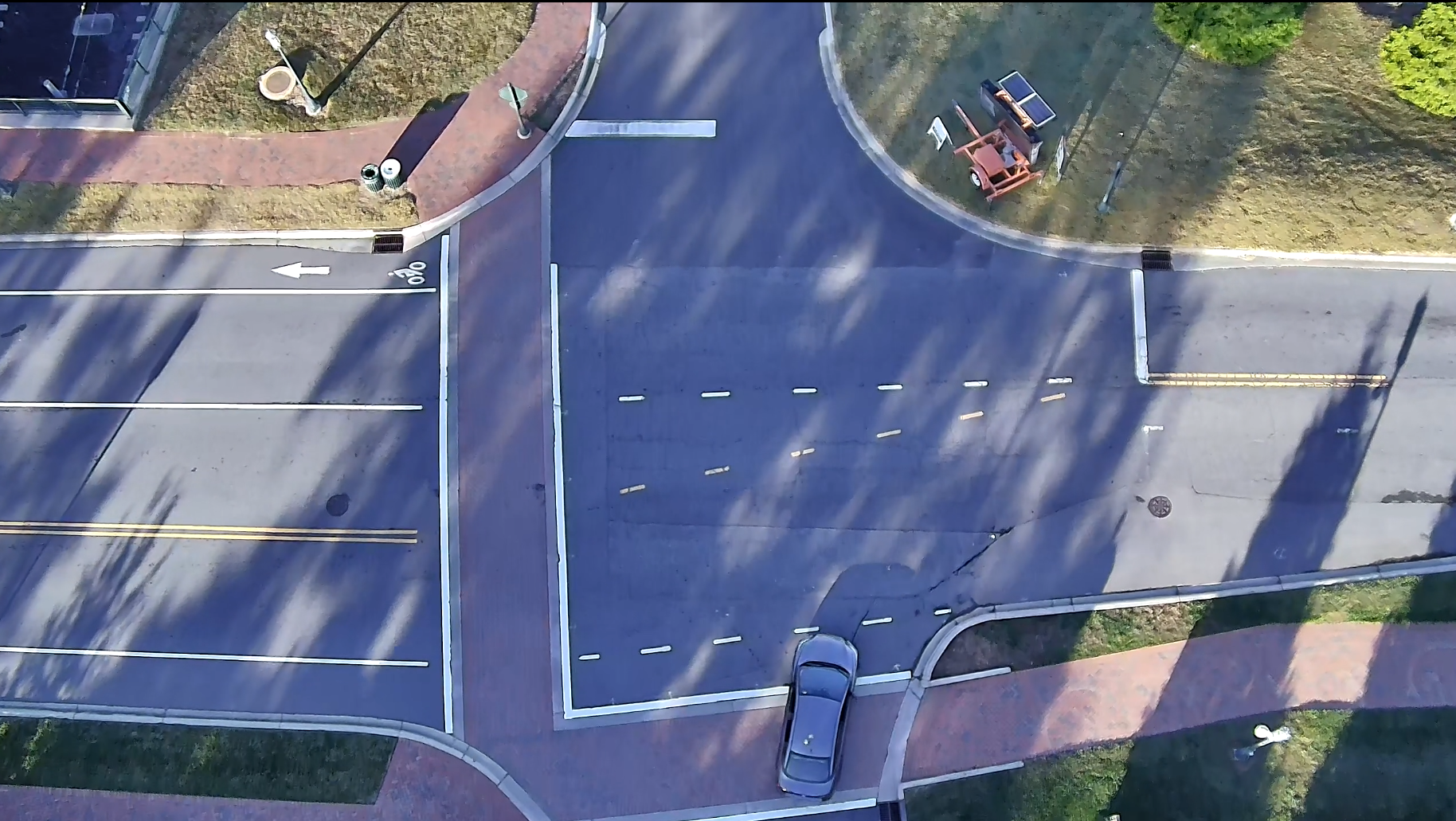}
    \caption{Example gimbaled camera data collected by the UNC Charlotte QAV 500 Drone while tracking a gray car along a straight road and through a parking lot.}
    %October 26, 2023 flight test. }
    \label{fig:imagery_oct26}
    \vspace{-1.5em}
\end{figure}

Vision-based perception, autonomy, and control have received great interest from both industry and academic researchers. Commercial platforms such as Skydio 2+ \cite{skydio2plus}, are capable of mapping and tracking and have been adopted in many real-world applications. Academic research groups have developed similar systems, often using fisheye camera configurations \cite{cloitre2019omnidirectional,gao2020autonomous}. For example, in \cite{cloitre2019omnidirectional}, fisheye cameras were mounted on a drone to allow omnidirectional sensing in an anti-polar orientation with spherical, cube, and/or capsule-type projections of fisheye images \cite{cloitre2019omnidirectional}. In \cite{kim2016landing}, a downward-facing fisheye camera was used to assist with landing on a moving platform. In other work, drones used vision-based feedback to follow people \cite{pestana2014computer,cheng2017autonomous} or dodge dynamic obstacles using event cameras \cite{sanket2020evdodgenet,falanga2020dynamic} or depth cameras \cite{tordesillas2022panther, zhou2022swarm}. However, few works have presented a UAV design capable of simultaneous urban mapping and tracking using open-source software and widely available low-cost components.

The contribution of this paper is the design and performance characterization of a quadrotor UAV with a unique imaging payload that supports research in mapping, hazard avoidance, and target tracking in urban environments (see Fig.~\ref{fig:imagery_oct26}). The platform is equipped with five cameras, including two pairs of fisheye stereo cameras that enable a nearly omnidirectional view (mapping and avoidance) and a two-axis gimbaled camera (target tracking). The platform is relatively small and low-cost and is designed from open-source and readily available components. The drone's performance is characterized in terms of acoustic noise, communication range, endurance, and top speed. The successful integration, functionality, and nominal performance of the data collection system is demonstrated through flight tests. The design and data presented may be useful to researchers developing platforms with similar capabilities and to facilitate planning urban flight experiments. 

The remainder of the paper is organized as follows. Section~\ref{sec:vehicle_design} describes the vehicle design and imaging payload integration. Section~\ref{sec:experiments} describes experimental results characterizing the performance, acoustic noise, communication range, speed, and mapping/tracking capabilities of the UAV. The paper is concluded in Sec.~\ref{sec:conclusion}.

\section{Vehicle Design}
\label{sec:vehicle_design}
This section describes the airframe design, hardware selection and integration of imaging sensors, and related image processing and control capabilities. 

% \begin{figure*}[t]
%     \centering
%     \includegraphics[width = 0.7\textwidth]{figures/updatedlabeledquad_v2.png}
%     \caption{The UNC Charlotte QAV500 drone}
%     \label{fig:QAV1}
% \end{figure*}

\begin{figure}[h!]
    \centering
    \includegraphics[width = 0.5\textwidth]{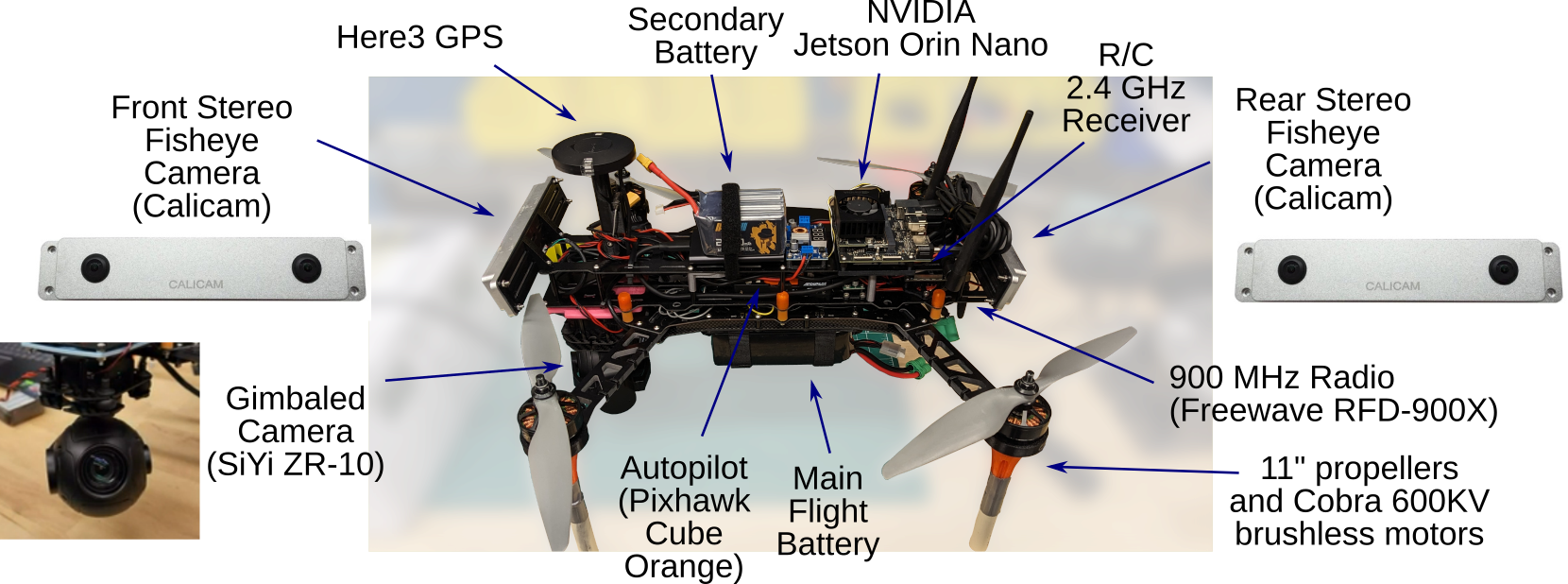}
    \caption{The UNC Charlotte QAV500 drone is equipped with a Cube Orange flight controller and carries a  NVIDIA Jetson Orin Nano, a  gimbaled camera, and two pairs of stereo fisheye cameras.}
    \label{fig:QAV1}
\end{figure}

\subsection{Airframe Design and Hardware}
The QAV500 is a custom-built quadrotor UAV that is developed from commercial/hobby-grade UAV components and in-house 3D printed parts (see Fig.~\ref{fig:QAV1} and Table~\ref{tab:components}). The main structure of the UAV is the QAV500 V2 H-style quadcopter frame made by Lumenier. The QAV500 V2 was chosen for its modular design, consisting of a rigid lower frame and an upper frame. Silicone bobbin dampers join the upper and lower frames, which isolate sensors from vibration. Various motor, propeller, and battery combinations were evaluated using xcopterCalc  \cite{ecalc} to maximize efficiency and endurance, the final selection is shown in Table~\ref{tab:components}. A 4in1 ESC was selected to simplify wiring. 
%Cobra CM-4008/24, 600KV brushless motors with 11in. propellers (APC 11x4.5MR-B4) and a Lumenier Elite 60A, 2-6S, 4in1 electronic speed controller (ESC) provide thrust to the UAV. 
A Cube Orange autopilot running Arducopter 4.4.0 firmware is used for flight control since it is a widely adopted and well-documented system.
The vehicle has a Here3 GPS with a companion Here+ RTK Base.  An NVIDIA Jetson Orin Nano with an attached WiFi board provides a companion computer intended for higher-level tasks such as sensor data acquisition and processing, state estimation/odometry, path planning, autonomy, guidance, navigation, and advanced control. A 900 MHz RFD-900X radio is used for telemetry/connection with a ground station computer running QGroundControl. The drone can be remotely piloted using a Spektrum 2.4 GHz transmitter.  
The main battery powers the flight controller, motors, radios, and GPS, and a secondary battery
%is an SMC Racing LiPo Battery (8000 mAh, 20C)  6S (6 cell) battery. A secondary Auline EX (2000 mAh, 120C) 6S battery 
powers the cameras and the companion computer through a step-down converter for an input of 19V.
\begin{table}[b]
\centering
\small
\caption{The main hardware components of the UAV design.}
\begin{tabular} { c | c  }
{\bf Component} & {\bf Model}  \\
\hline
Frame & Lumenier QAV500 V2  \\
Frame Arms & Lumenier QAV 500mm Size G10  \\
Motors & Cobra CM4008/24, 600KV  \\
Propellers & APC 11x4.5MR-B4   \\
ESC& Lumenier Elite 60A, 2-6S, 4in1 \\
Flight Controller & Cube Orange ADS-B Board \\
Main Battery & SMC Racing (LiPo, 8000 mAh, 20C, 6S)\\
Secondary Battery & Auline EX (2000 mAh, 120C, 6S) \\
GPS & Here3 GPS module  \\
Radio Receiver & Spektrum SPM9745 DSMX Receiver \\
Gimbaled Camera & SiYi ZR-10\\
Fisheye Cameras & Calicam \\
Computer & NVIDIA Jetson Orin Nano \\
Telemetry Radio & RFD-900X (900 MHz)
\end{tabular}
\label{tab:components}
\end{table}

The landing gear, frame edge protectors, and electronics mounts were designed using a CAD model of the UAV frame. The landing gear was printed out of PETG filament and extended with 0.5~in diameter, 3~in long PVC pipe segments press-fit into the printed landing gear.  The motors and landing gear are mounted concentrically using the same bolts.  Frame arm edge protectors, printed using FLEX filament, are installed under the motors.  The battery mount is printed using FLEX filament and serves to prevent the battery from sliding during flight and to protect the battery from exposed screws. The GPS mount is 3D printed out of PETG plastic and consists of a clamping swivel and a tower. Double-sided adhesive tape secures the GPS to the tower. The tower then connects to a clamping swivel with detents that allow the GPS to be locked into an upright position for use and a forward-facing lowered position for storage, allowing multiple UAVs to be stacked.

The weight of the assembled platform, with minimum instrumentation required for flight, is 2.8~kg and has a diagonal dimension of 50~cm.  The endurance and range of the platform were predicted using the xcopterCalc \cite{ecalc} performance calculator (Table~\ref{tab:ecalc}) both with and without the weight and current draw of optional vehicle payloads (including cameras, NVIDIA computer, and secondary batteries).  The exact model of the ESC was not available in the xcopterCalc database and an approximated model was used instead.  The ESC parameters were set to 60~A for continuous current capacity, 100~A for maximum current capacity, a general resistance of 0.0025~$ \Omega$, and a weight of 20~g, per the manufacturer's specifications.  Models of all other chosen parts were available.  Approximated values of flight endurance and range assuming a standard model of drag at various airspeeds are shown in Table~\ref{tab:ecalc}.  No limit was set for the maximum vehicle tilt allowed by the flight controller in this analysis. The fully instrumeted UAV weighs 4.1~kg and costs approximately \$1120. 

% Modified table setup based on https://www.tablesgenerator.com/#
\begin{table}[b]
\centering
\small
    \caption{Estimated performance characteristics of the UAV using \cite{ecalc}. The platform with the imaging payload has an additional current draw of 3A and an additional weight of 1350 grams.}
    \begin{tabular}{p{0.5in}|p{0.6in} | p{0.4in}|p{0.6in} | p{0.4in}}

     & \multicolumn{2}{|c|}{ {\bf Base Platform} } &  \multicolumn{2}{|c}{ {\bf With Payload} }  \\
     \hline
     {\bf Airspeed} \newline (km/h)  & {\bf Endurance} (sec) & {\bf Range} (m) & {\bf Endurance} (sec) & {\bf Range} (m) \\
    \hline
    0    & 1122 & 0 & 618 & 0 \\
    10  & 1080 & 3000 & 600 & 1667 \\
    20   & 900 & 5000 & 530 & 3000 \\
    30   & 825 & 5700 & 440 & 3667 \\
    35.5 & 570 & 5650 & 385 & 3834 \\
    40   & 500 & 5500 & -- & -- \\
    46   & 400 & 5200 & -- & -- \\ 
    \end{tabular}%
    \label{tab:ecalc}
\end{table}

\subsection{Target Tracking Sensor and Autonomous Behavior}
%For target tracking mapping/avoidance tasks, respectively, the SiYi ZR10 gimbaled  camera and CaliCam fisheye stereo cameras were selected and integrated onto the platform.
% \begin{figure}[h!]
%     \centering
%     \includegraphics[width=3in]
%     {figures/monoscopic_gimbal_arsl.png} \\
%     \includegraphics[width=1\linewidth]{figures/dual_fisheye_stereo_arslquad.png}
%     \caption{The hardware and mounting of the gimbaled camera and dual stereo fisheye sensors for the UAV platform.}
%     \label{fig:monoscopic_gimbal_camera}    
% \end{figure}

This SiYi ZR10 Gimbal Camera is a low-cost gimbaled camera that supports two-axis (pan/tilt) motion and can be controlled directly using MAVLINK commands to point at a desired latitude and longitude. A two-position switch on the radio controller starts/stops the recording. The video feed is recorded to an onboard micro SD card or can be streamed using ethernet to the NVIDIA Orin via RTSP. The sensors connected to the autopilot (accelerometers, gyros, compasses, and GPS) are logged on the autopilot's black-box SD card. 

Target tracking requirements are often specified by altitude and standoff distance relative to the target. The standoff distance can be achieved in many ways, for example, by continuously orbiting a target, or tracking it from a particular orientation relative to the target velocity (i.e., so the target appears in the same orientation on the video feed). Perhaps the simplest tracking approach, and the one used in this work, is to hold a fixed relative position to the target. 
%or The UAV can follow the target either in a {fixed-track} orientation (where it always views the target from the same relative position) or from a {motion track} geometry where it maintains a particular orientation, such as perpendicular, relative to the {velocity} of the target.  The drawback of the motion track approach is that it can require aggressive maneuvering during more complex target motions. 
To command a constant relative position that is directly overtop of the target, we utilize dronekit \cite{dronekit} to send MAVLINK messages. A latitude/longitude coordinate of a moving ground target is set as the desired waypoint for the UAV and the desired target pointing position for the gimbal. 
%to command the UAV to move to a prescribed standoff distance (fixed-track) from a target that is transmitting its GPS position via radio . 
Presently, the target latitude/longitude is directly provided by the target itself (transmitted over the 900 MHz radio). This provides an automated capability to collect visual tracking data and test algorithms using perfect knowledge of the target's position (as an example, see Fig.~\ref{fig:imagery_oct26}). In future work, this will be replaced with an online estimate of the target position derived from onboard imaging sensors. 

\subsection{Mapping Sensors and Related Perception Algorithms}
\begin{figure}[h!]
    \centering
    \includegraphics[width=.35\textwidth]{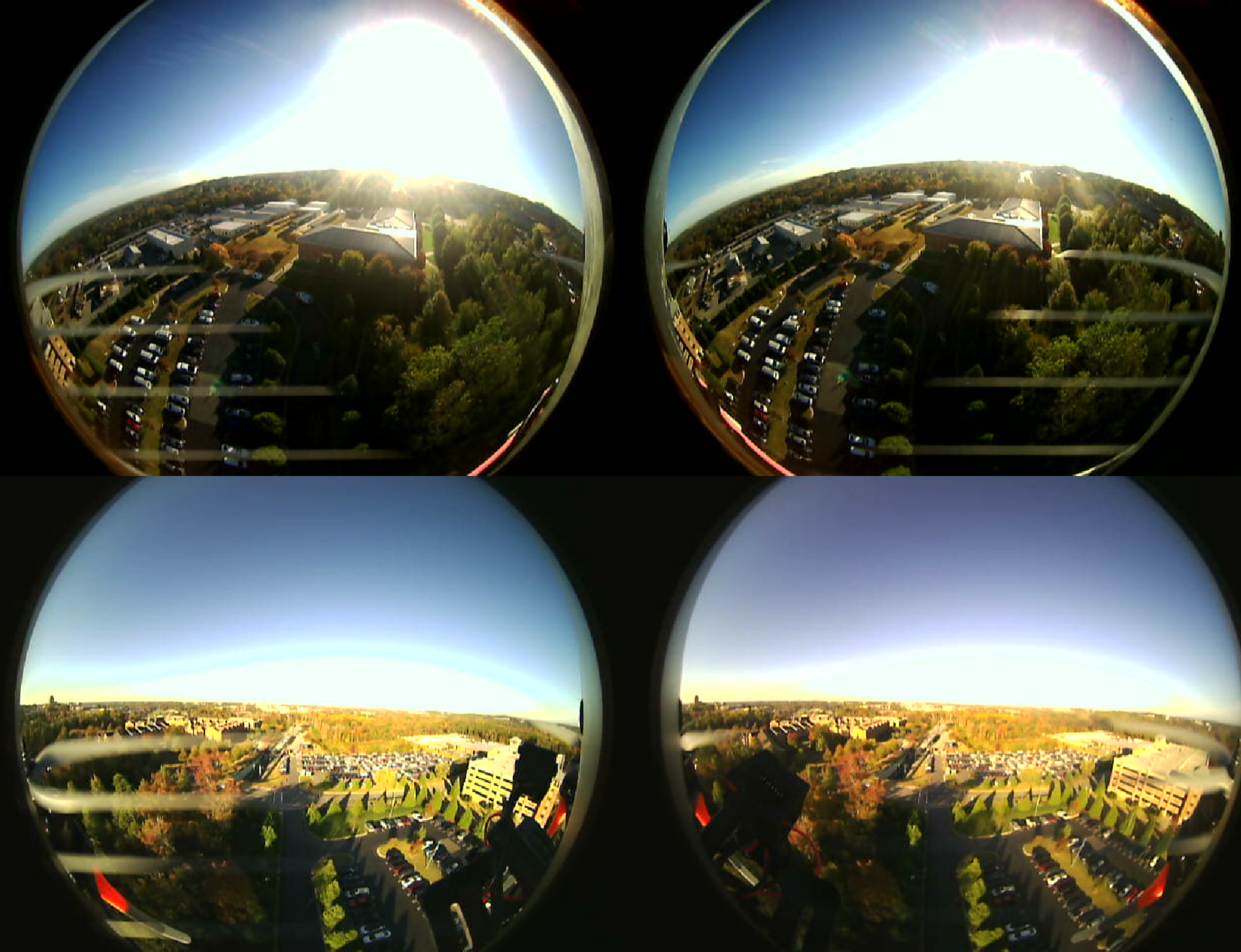}
    \caption{The front (top) and rear (bottom) stereo fisheye images during the tracking test. The horizontal aberrations are due to the presence of the propeller in the field of view.}
    %\caption{A comparison of two different resolutions for the stereo fisheye cameras. The top image is full resolution 2560 x 960 px while the bottom image is 25\% resolution 1280 x 480 px.}
    \label{fig:calicam}
\end{figure}

The Calicam \cite{CaliCam} fisheye stereo cameras were chosen as the perception sensor used for mapping and  hazard detection.
The CaliCam is an electronic rolling shutter camera that has a 12~cm baseline, a 200$^\circ$ horizontal field of view, a pixel size of 3.75 x 3.75~$\mu$m, and a resolution of 2560 x 960 at 30 Hz. The wide field of view, precise factory calibration, and compact and low-cost design made it an ideal choice.
% are as follows:
% \begin{itemize}
%     \item Baseline: 12 cm
%     \item Resolutions: 2560 X 960 @ 30Hz, 2560 X 720 @ 30Hz, 1280 X 480 @ 30Hz
%     \item Field of View: Horizontal 200°
%     \item Pixel Size: 3.75 X 3.75 µm
%     \item Scanning Mode: Electronic Rolling Shutter
% \end{itemize}
The cameras were mounted on the front and rear of the UAV using PETG mounts that were inclined downward at $15^{\circ}$ angles. 
This arrangement provides a nearly omnidirectional view that minimizes blind spots. 
%For autonomous drones, this means a more complete understanding of their environment, leading to better decision-making and safer navigation. The configuration of a stereo camera pair that minimizes the computational load and simplifies calculations is often referred to as a "baseline parallel" or "long baseline" setup. In this configuration, the two cameras are positioned parallel to each other with a fixed horizontal separation (baseline) and they are pointed straight ahead. 
The cameras are designed with a long baseline configuration, wherein the two cameras are positioned parallel to each other with an equivalent pointing direction and fixed horizontal separation, simplifying mapping tasks such as triangulation and disparity calculations.
%The advantages of a baseline parallel configuration include a simplified epipolar geometry, straightforward disparity calculation, reduced rectification complexity, and easier calibration.
An example of the CaliCam imagery is shown in Fig.~\ref{fig:calicam}. 
While the fisheye cameras are the main mapping sensors, the gimbaled camera can also be used for mapping. 
%The two images were taken with different resolutions which allows for more or less information to be captured by the stereo camera. 

\begin{figure}[h!]
    \centering
    \includegraphics[width=.4\textwidth]{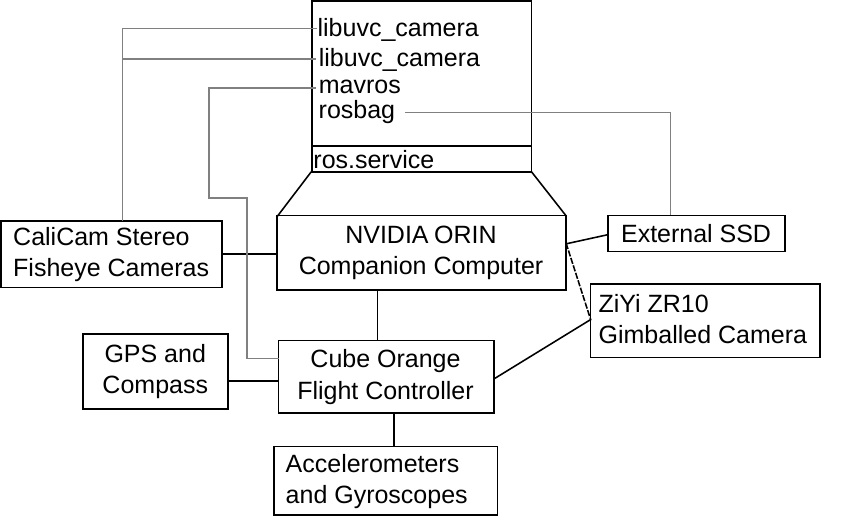}
    \caption{A diagram of the hardware connections and ROS nodes running on the NVIDIA Orin, the gray lines are software connections, the black lines are the hardware connections, and the dashed black line is a future hardware connection.}
    \label{fig:ros}
\end{figure}
\subsection{Data Collection Computer}
%% ros setup description
The UAV is equipped with an NVIDIA Orin computer that can be used for data collection, and running autonomy and perception algorithms. The Pixhawk Cube autopilot is connected to the NVIDIA Orin via a UART connection, and the CaliCams are connected to the NVIDIA Orin via high-speed USB 2.0 cables. To synchronize all of the data streams, the NVIDIA Orin uses the Robot Operating System (ROS) to synchronize the data streams' timestamps. The three main ROS nodes running on the NVIDIA Orin are MAVROS---an intermediary that converts MAVLink messages to ROS messages---and two libuvc\_camera nodes---a node that connects to most USB cameras. To record the desired data streams, the \texttt{rosbag} command writes ROS topics to a drive. The amount of information coming from the CaliCams with full resolution was too high for the USB-C solid-state hard drive to write, so the CaliCams' resolution was reduced from 2560 x 960 to 1280 x 480 which allowed for all of the data to be captured.
To start the ROS system two Linux services were created, one as the super user to modify permissions for the CaliCam USB ports and run any startup tasks, and a second as a regular user to call the ROS launch file to start the ROS nodes. These services automatically start while the NVIDIA Orin is booting. For a diagram of the hardware connections and ROS nodes refer to Fig.~\ref{fig:ros}.

%For the high-resolution image in Fig.~\ref{fig:calicam}, a DJI FPV quadrotor is equipped with an NVIDIA Orin computer and connected to the CaliCam fisheye camera, recording throughput must be maximized to get good FPS. To do this, the NVIDIA Orin is set to "multi-user" mode to start with a command-line interface instead of the graphical interface to preserve resources and prevent any background graphical tasks. Next, a startup script was created that when the Orin is powered on it will create a temporary ramdisk, create a unique folder to store the fisheye images in, and start a Python script to begin recording. The reason for the ramdisk being created is to increase throughput on the recording as it will be saved immediately to RAM which is known to be significantly faster than saving to disk. As for the Python script, it is designed to be multi-threaded where one thread records images from the CaliCam to the ramdisk while another thread calls the command \texttt{rsync} periodically that will copy the contents of the ramdisk to the unique folder and then delete the contents once the transfer is complete.

\section{Experimental Characterization of Quadrotor Capabilities}
\label{sec:experiments}
\subsection{Quadrotor Performance Characterization}
\begin{figure}[h!]
    \centering    \includegraphics[width=0.35\textwidth]{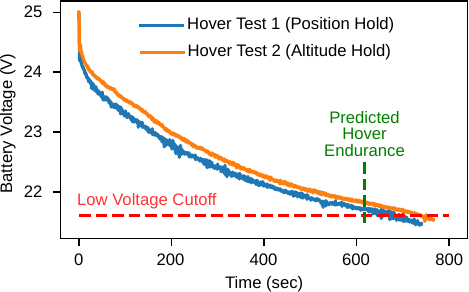}
    \caption{Battery voltage while the UAV hovers in place. The blue line was a hover test with GPS hold and manual altitude control with a flight time of 11min 27s, while the orange line was a hover test with GPS hold and automatic altitude control with a flight time of 12min 24s. The low voltage cutoff was 21.6V corresponding to 3.6V per cell. The green line is the predicted flight time of 10min 18s.}
    \label{fig:battery}
\end{figure}

Two hover tests were performed to determine the safe hovering time of the UAV, which was determined to be about 11~min based on the battery voltage drop observed in Fig.~\ref{fig:battery} (the flight battery cutoff is a 3.6V cell average). In the first trial, a pilot held a constant altitude while in position hold flight mode (the UAV maintains a fixed horizontal position and the pilot controls the altitude). In the second trial, an identical battery was used and the UAV was in altitude hold flight mode with a target altitude of 2~m. The UAV's altitude varied by approximately 1.5~m during the automated altitude hold test and by less than 0.5~m in the position hold test.

\begin{figure}[h!]
    \centering
    \includegraphics[width=.4\textwidth]{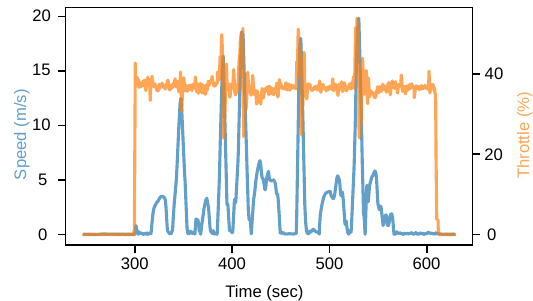}
    \caption{Horizontal speed (blue) and throttle (orange) during high-speed testing experiment. The maximum speed was 19.8~m/s.}
    %The maximum speed obtained was $19.8$~m/s and the flight was at the limits of our testing area.}
    \label{fig:speed}
\end{figure}

Another experiment was carried out to determine the maximum speed of the UAV. For this test, the maximum pitch/roll angle in the autopilot settings was changed to
45$^{\circ}$ and the vehicle was flown manually in a straight line. Fig.~\ref{fig:speed} shows a plot of the throttle and horizontal speed. Due to constraints in the test area, the throttle was limited to less than 50\%, and at this throttle setting a speed of 19.8~m/s was achieved. Other performance metrics for the QAV are estimated by the xcopterCalc \cite{ecalc} and shown in Table~\ref{table:ecalc_performance_estimates}.

\begin{table}[h]
\small
    \begin{center}
        \caption{UNC Charlotte UAV performance estimates from \cite{ecalc}}
        \begin{tabular}{ l | c }
        \label{table:ecalc_performance_estimates}
        {\bf Parameter } & {\bf Value } \\ \hline
            Hover flight time & 10.4 min.\\
            Thrust-weight ratio & 1.7:1\\
            Specific thrust & 5.23 g/W\\
            Per motor current draw at hover & 9.11 A\\
            Per motor power consumption at hover & 196 W\\
            Total power consumption at hover & 676.7 W\\
            Efficiency at hover & 83.7 \%\\
            Linear throttle position at hover & 66 \%\\
            Per motor current draw at max throttle & 20.77 A\\
            Per motor power consumption at max throttle & 429.9 W\\
            Total power consumption at max throttle & 1475.2 W\\
            Efficiency at maximum throttle & 80 \%\\
            Maximum speed & 18.06 m/s\\
            Maximum rate of climb & 6 m/s\\
        \end{tabular}
    \end{center}
\end{table}

\begin{figure*}[t]
\vspace{.02in}
\centering
\includegraphics[width=0.75\textwidth]{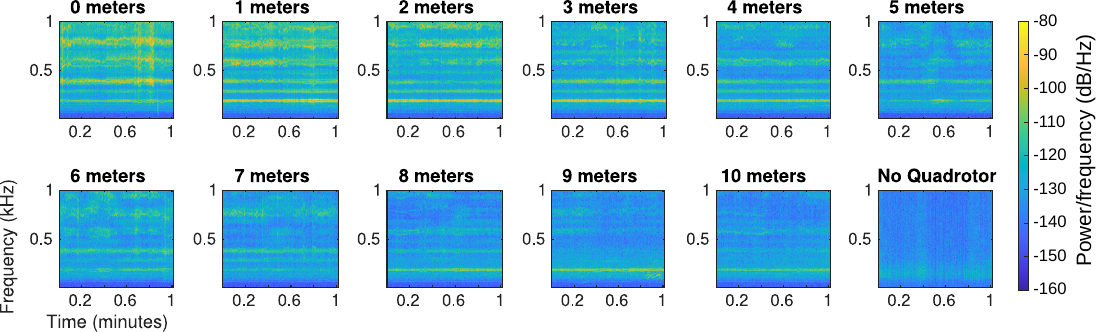}
\caption{Acoustic data recorded during hovering flight test. The UAV's position was varied from 0 to 10 meters. The yellow color indicates higher power at the specific frequency while the blue color indicates lower power.}
\label{fig:acoustics}
\end{figure*}

\subsection{Quadrotor Noise Characterization}
A UAV operating in an urban environment at a relatively low altitude generates significant noise that can be undesirable to bystanders. Characterizing the baseline noise generated by the UNC Charlotte QAV500 drone provides information to model the observed level of the noise on the ground at different UAV altitudes. An experiment was conducted on the UNC Charlotte campus with a UAV in a fixed hovering position at 5~m altitude. One-minute length audio recordings were collected from zero to ten meters in one-meter intervals by moving a tripod with a mounted Dayton Audio UMM-6 microphone (frequency response 20~kHz). Onboard telemetry recorded drone states, including altitude, position, and motor RPM. A control data set (with no UAV) and a ground-level data set were also recorded. Spectrograms visually representing the energy content of frequencies in the 10 Hz-1kHz range over time as shown in Fig.~\ref{fig:acoustics}.  The sound intensity of the UAV decreases with horizontal distance. The spectrogram displays a pattern of high energy frequency bands corresponding to the motor RPM, blade frequency, and their harmonics.
\subsection{Quadrotor Communication Range Characterization}
\begin{figure}[h!]
\centering
\includegraphics[width=0.45\textwidth]{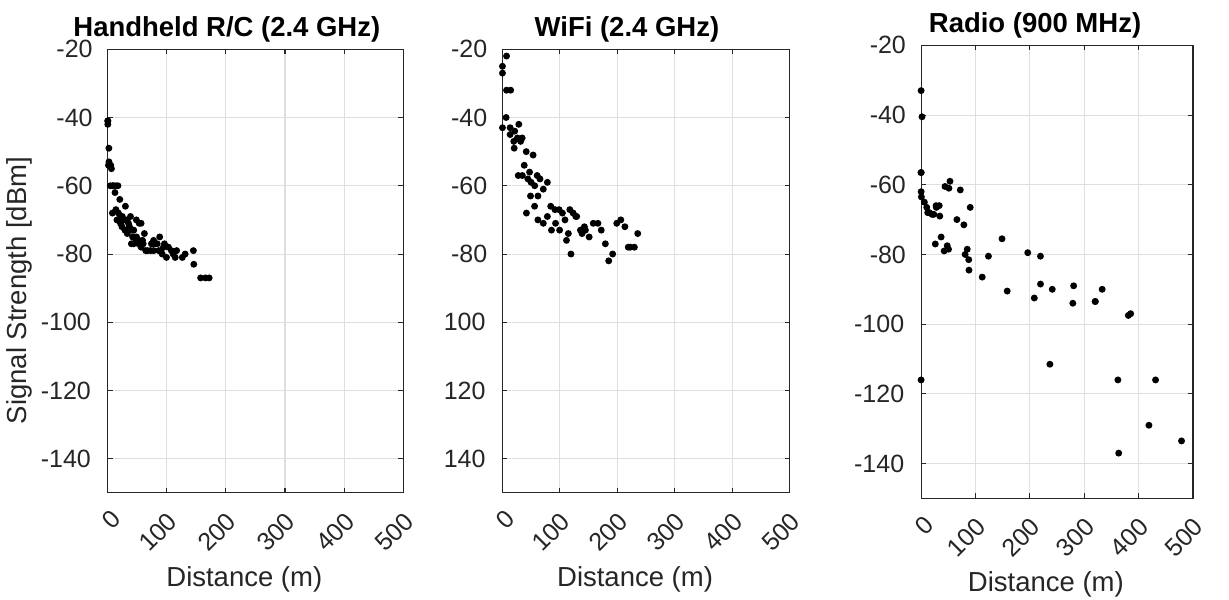}
\caption{Communication range characterization. A 2.4GHz radio was compared with 2.4GHz WiFi and a 900MHz radio. The 2.4GHz communication methods lost connection around 200m to 300m while the 900MHz radio held connection up to 500m.}
\label{fig:comms}
\end{figure}

The UAV can potentially utilize three forms of communication with the ground station operators (WiFi, 900 MHz radio, and 2.4 GHz transmitter). Communication interruptions can lead to loss of vehicle control and characterizing the range of the communication systems through testing allows for increased safety and more reliable mission planning.
Range testing was conducted on UNC Charlotte’s campus with all three of the aforementioned communication channels while recording RSSI (dBm) and GPS data. 
Figure~\ref{fig:comms} illustrates the change in radio strength over distance during these tests. Note that the WiFi module used in this experiment was comparable to but not the same as the one used on the NVIDIA Jetson Orin Nano. 
The graphs show that a range of $-40$ dBm to $-90$ dBm for the transmitter, $-20$ dBm to $-80$ dBm for the WiFi, and $-30$ dBm to $-120$ dBm for the radio modem is suitable for reliable communication. The radio modem has the longest range of communication. A sharp decrease in signal strength is evident across all three communication modes during the beginning of the experiment, but the decrease levels out at the end of the experiment.
At around -70 to -80 dBm range, the signal strength became approximately constant for the R/C transmitter. Longer-range pilot transmitters and elevated antenna poles are being considered for future operations. 
\subsection{Mapping and Tracking Flight Demonstration}
\label{sec:results}
\begin{figure*}[t]
    \vspace{.02in}
    \centering
    \includegraphics[width=0.7\linewidth]{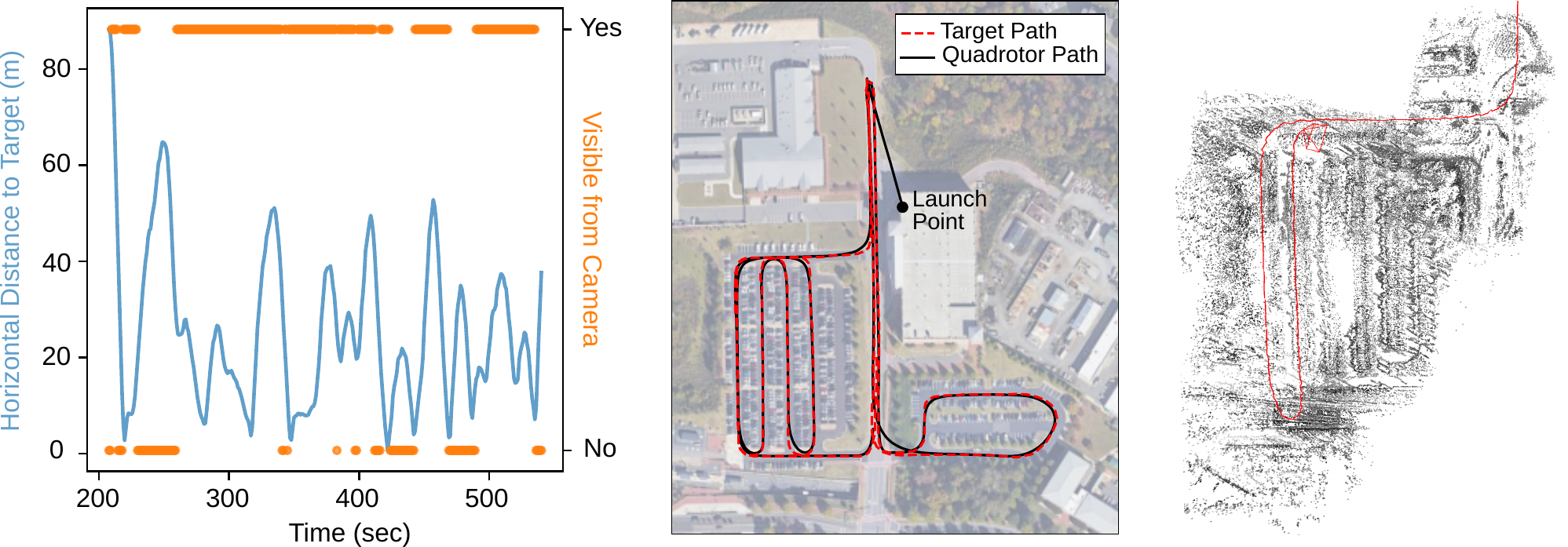}    
    \caption{Overview of tracking experiment. Left: A plot of the horizontal distance from the UAV to the target throughout the tracking mission overlayed with a graph of when the target was visible in the UAV's gimballed camera. Middle: A GPS trace of the target motion (red) and the actual UAV path (black). 
    %Right: Mapping results of a video segment of the flight tests using DSO. Left: Satellite image from Google Maps of a section of the testing area. 
Coll    Right: A point cloud that shows the structure of the parking lot (constructed from images like Fig. \ref{fig:imagery_oct26}). The red curve shows the UAV's odometry and the pyramid at the end of the trajectory depicts the current pose of the gimbaled camera.}.
    \label{fig:tracking_graphs}
    \vspace{-2em}
\end{figure*}

A flight test was conducted on October 26, 2023 on UNC Charlotte's campus to demonstrate the target tracking and mapping capabilities of the UAV. The test was conducted from the roof deck of a five-story parking garage at  Poplar Lane, Charlotte, NC (latitude/longitude: 35.31389$^\circ$, -80.73175$^\circ$). 
\begin{figure}[h!]
\centering
\includegraphics[width = 0.48\textwidth]{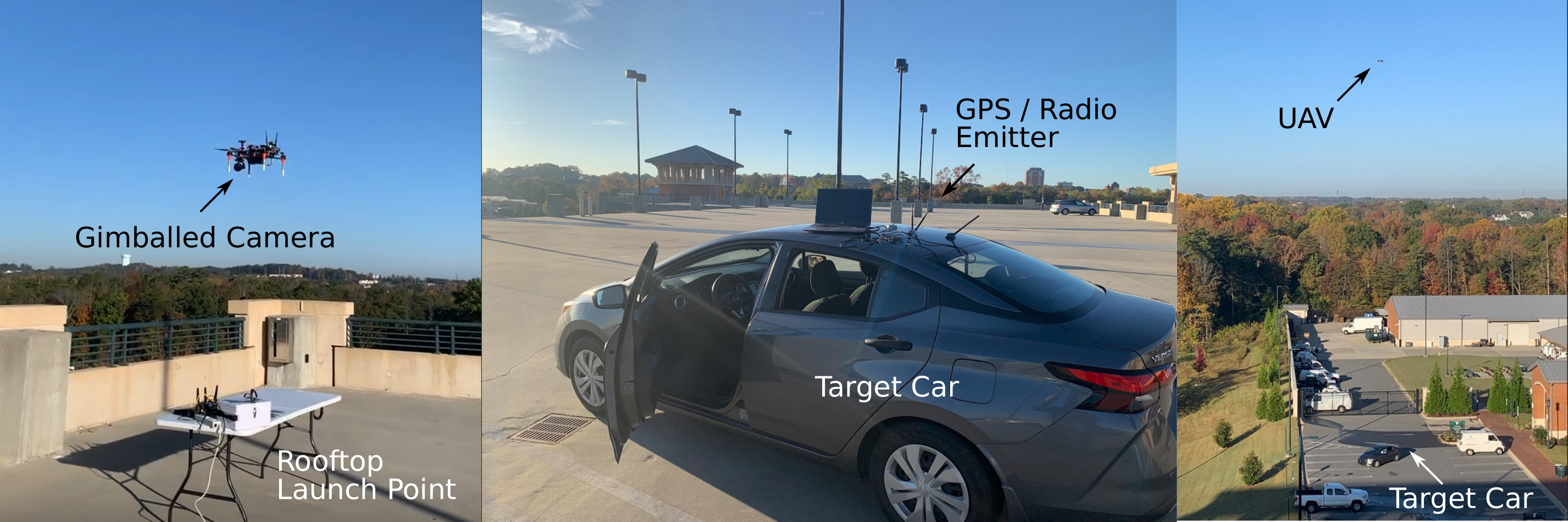}
    \caption{The setup of the tracking experiment included  a GPS-tagged target car transmitting its position and a  UAV launched from the roof of a  parking deck. The UAV took off to the target altitude and then flew to its starting position directly above the car. The UAV pilot monitored the flight from atop the parking deck.}
    \label{fig:setup_oct26}
\end{figure}
This location provides the pilot with a clear view of the UAV during tests. The flight test was designed to test the target tracking capability as well as fisheye imagery collection and mapping.  The experiment setup is shown in Fig.~\ref{fig:setup_oct26}.  A target vehicle was outfitted with a roof-mounted 900 MHz antenna, GPS unit, and laptop, to broadcast UAV desired waypoints while in motion. After launch, the UAV was commanded to continuously follow directly over the top of the vehicle and point the gimbal at the vehicle location. 
%The UAV was launched from the rooftop of a parking garage on UNC Charlotte's campus. Simultaneously a target car with a  roof-mounted GPS/radio emitter was positioned at a parking lot at ground level.  
Once the UAV was hovering above its launch point the car initiated transmission of drone waypoints and gimbal pointing coordinates read from the car-mounted GPS unit.
The UAV tracked the target for approximately 5.5 minutes. Example imagery collected is shown in Fig.~\ref{fig:imagery_oct26}. The UAV had a maximum velocity constraint set at 8~m/s and the altitude corresponded to a standoff distance of approximately 55 meters. The UAV tracked the car through two different parking lots and along the road, Fig.~\ref{fig:tracking_graphs}. The UAV stayed within 60 meters of the target car during the test and kept the target in the gimbal camera image 72.3\% of the time from manual video analysis.
%of the resulting video.

% \begin{figure}
%     \centering
%     \includegraphics[width=.35\textwidth]{figures/tracking_distance.pdf}
%     \includegraphics[width=0.55\linewidth]{figures/tracking_gps_v2.pdf}    
%     \caption{Top: A plot of the horizontal distance from UAV to the target throughout the tracking mission overlayed with a graph of when the target was visible in the UAV's gimballed camera. Bottom: A GPS trace of the target motion (dashed red line) and the actual UAV path (black line)}.
%     \label{fig:tracking_graphs}
% \end{figure}

Follow-on work should also consider adding additional control logic to keep track of how close the gimbal is to its mechanical limits and perform maneuvers to avoid hitting this limit (causing the gimbal to ``unwind'').  Another question to be explored in future work is the joint optimization of vehicle motion and gimbal motion---that is, when should the vehicle move to improve the view of the target versus rotate the gimbal and/or how to perform this maneuver optimally.
%The experiment was restarted several times due to communication issues previously not encountered during shorter range tests. Communication issues (disconnection/interruption) occured on the 900 MHz channel using the Freewave radio unit (for passing GPS coordinates and commands) as well as on the manual 2.4 GHz controller. Future testing should utilize longer range equipment, for example the Express LRS system for manual control. 

% \subsection{Mapping Demonstration: Flight Tests: October 26, 2023}
% From the same location on the rooftop the UNC Charlotte QAV platform was deployed to follow a survey pattern in an industrial area. The total flight time was about 4 minutes.
%\begin{figure}[h!]
%    \centering
%    \includegraphics[width = 0.4\textwidth]{figures/industrial_area.png}
%    \caption{Industrial area used to collect fisheye camera data.}
%    \label{fig:industrial}
%\end{figure}
%Details of the survey pattern and example telemetry, including speed in the north, east, down directions $(x,y,z)$ is shown in Fig.~\ref{fig:industrial}.
% \begin{figure}[h!]
%     \centering
%    % \includegraphics[height = 1.7 in]{figures/3dQuadrotorPlot.pdf}
%     \includegraphics[width = 1.5 in]{figures/QuadrotorFlight.pdf}
%     \includegraphics[width = 2.5 in]{figures/QuadrotorVelocity.pdf}
%     \caption{A plot of the 3D trajectory during a waypoint mission. A 2D plot of the trajectory during the same waypoint mission overlayed on a map. The velocities of the UAV during the waypoint mission.}
%     \label{fig:industrial}
% \end{figure}

{Fig. \ref{fig:tracking_graphs} illustrates the mapping capability of the UAV by applying the state-of-the-art DSO (Direct Sparse Odometry) algorithm \cite{engel2017direct} to the imagery collected during the experiment. DSO is a visual odometry technique that adapts Structure-from-Motion (SfM) methods for 3D reconstruction. It estimates the camera motion and the sparse 3D structure of the environment from a sequence of 2D images by minimizing photometric errors. Our implementation adhered to the original configuration \cite{engel2017direct} with the default parameters. The point cloud visualizes the parking lot's structure, while the red curve represents UAV odometry, and the pyramid symbolizes the current pose of the gimbaled camera. It is important to note that DSO is not optimized for cameras with frequent and constant pose changes, such as gimbaled cameras. Consequently, achieving highly optimized reconstructions directly from gimbaled camera data poses challenges. This experiment demonstrates the mapping potential of our design and future work will utilize the stereo fisheye imagery for mapping.}

% \begin{figure}
%     \centering
%     \includegraphics[width=0.95\linewidth]{figures/dso_sat}
%     \caption{Mapping results of a video segment of the flight tests using DSO. Left: Satellite image from Google Maps of a section of the testing area. Right: A point cloud that shows the structure of the parking lot (constructed from images as in Fig. \ref{fig:imagery_oct26}). The red curve shows the odometry of the UAV and the pyramid at the end of the trajectory depicts the current pose of the gimbaled camera.}
%     \label{fig:DSO_mapping}
% \end{figure}

\section{Conclusion}
\label{sec:conclusion}
This paper presented the design of a quadrotor UAV that is tailored for conducting research related to autonomous mapping, hazard avoidance, and target tracking. The platform has a unique sensor arrangement that includes 5 imaging sensors (two pairs of stereo fisheye cameras and a gimbaled camera). Various characteristics of the UAV were experimentally identified, including its endurance (11 minutes), range (3.8 km), and top speed (20 m/s). The noise level of the UAV as a function of distance was also evaluated from 0 to 10 m. The practical communication range of the onboard radios and WiFi, in the presence of obstructing buildings, was also determined. Lastly, the capabilities of the UAV to perform mapping and tracking tasks were demonstrated through a flight experiment wherein the UAV followed a GPS-tagged vehicle moving through a parking lot, and data collected was used to construct a map of the scene. The information reported may be useful to researchers developing similar systems. 

In future work, we aim to enhance  the UAV's software architecture to support onboard testing of autonomous path planning, control, and perception algorithms. The platform will be used to support planned work involving tracking moving ground targets in the presence of occlusion and the use of fisheye cameras to support hazard avoidance. 

\section{Acknowledgments}
We acknowledge the help of Jacob Harrison and John Driver in collecting acoustic and communication range data, Gunner Petrea with development of the quadrotor airframe, and Joey Philips and Jim Conrad in assisting with the gimbaled camera and experimentation. 

\bibliographystyle{IEEEtran}
\bibliography{main}
\end{document}